\documentclass[conference]{IEEEtran}
\IEEEoverridecommandlockouts
\usepackage{amsmath,amssymb,booktabs,array}
\usepackage{xcolor}
\definecolor{navy}{RGB}{0,102,255}
\usepackage{balance}
\usepackage{pgfplots}
\usepackage[hidelinks]{hyperref}
\pgfplotsset{compat=1.18}
\newcommand{\Gtw}{G26}
\newcommand{\Gth}{G31}
\newcommand{\Nem}{N120}
\newcommand{\Qwe}{Q35}
\newcommand{\Oss}{O120}
\newcommand{\Qfl}{Q38}
\newcommand{\ind}{\mathbf{1}}
\newtheorem{lemma}{Lemma}
\newtheorem{corollary}{Corollary}
\usepackage{eso-pic}
\newcommand{\arxivnotice}{\AddToShipoutPictureFG*{%
  \AtPageLowerLeft{\put(\LenToUnit{0.5\paperwidth},\LenToUnit{0.32in}){%
    \makebox[0pt]{\parbox{\textwidth}{\centering\scriptsize \copyright\ 2026 IEEE. Personal use of this material is permitted. Permission from IEEE must be obtained for all other uses, in any current or future media, including reprinting/republishing this material for advertising or promotional purposes, creating new collective works, for resale or redistribution to servers or lists, or reuse of any copyrighted component of this work in other works.}}}}}}
\begin{document}
\arxivnotice
\title{When Multi-LLM Agreement Misses Shared Errors:\\Evidence from Debt Covenants in 10-K Filings\thanks{This work was supported by the PNC Presidential Fellowship Research Award that I hold, which funded much of the computing used here. The views expressed are my own and not necessarily those of PNC. Parts of this research ran on ORCHARD, a high-performance cloud computing cluster, and I thank Carnegie Mellon University for making it available to me.}}
\author{\IEEEauthorblockN{A. Orhun Gun}
\IEEEauthorblockA{\textit{Accounting AI Research Lab} \\
\textit{Carnegie Mellon University} \\
Pittsburgh, PA, USA \\
orhun@cmu.edu}
}
\maketitle
\begin{abstract}
To what degree does the convergence of two different LLMs on the same output signal accuracy? Multi-LLM consensus is widely used to automate data annotation and extraction tasks. This usually rests on the silent assumption that the models are independent, meaning that their errors are orthogonal. On this premise, agreement is taken to certify accuracy, and only the absence of agreement calls for a deeper review, by a larger model or a person. In this paper I study this fundamental assumption and show that raw agreement can hide an error that a pipeline rule induces in every model. I test this on a debt covenant extraction pipeline I built for accounting research. Four open-weight models read the same selected text from a stratified sample of 15,600 SEC 10-K filings under the same rules. I change one rule they all share: whether extractions from debt exhibits count alongside the 10-K body. With exhibits counted, every model's covenant measure rises, and the measured growth in covenant threshold disclosure since the early 2000s shrinks by about two fifths. Yet disagreement among the models barely moves. Reading sampled filings in full with separate LLMs, I find that about one in twenty filings that all four models label negative still states a qualifying threshold. All 16 sampled misses trace to rules the models share, such as which text they read or which clauses the prompt asks for. A review process triggered only by disagreement would miss these shared omissions.
\end{abstract}

\section{Introduction}
If four people could all testify to the same event, we would find that more credible than a single witness.  The agreement of the four witnesses is persuasive because each saw the event independently of the others. It is worth far less if all four watched through the same window, because then they not only see the same things but are equally blind to whatever that window hides. Agreement among large language models (LLMs) is often read the first way: if four different models return the same answer, the answer seems to belong to the document, not to the reader. To reduce human review, and therefore increase labor efficiency, a cross-model consensus workflow accepts labels that LLMs agree on and sends disagreements to a person~\cite{yuan}. A production fact-extraction pipeline can then go further and treat agreement as proof of accuracy, so reviewers never reexamine the agreed answers. Yet LLMs may share more of the window than this reading assumes. Errors are shown to be strongly correlated across LLMs, and more capable models err more alike~\cite{kim,goel}. Models in one pipeline share it literally: they read the same selected passages under the same prompt and rules.

This paper studies the agreement statistic itself. It asks what agreement may be silently hiding when the object is a financial research variable, rather than a model response. LLM outputs increasingly enter accounting and finance research as data~\cite{dekok,ludwig}. Generated variables carry their measurement error into downstream regressions~\cite{battaglia,egami}. A debt study, for example, may need an indicator for a disclosed covenant threshold. Such a threshold is often a tripwire whose violation shifts control rights to creditors~\cite{chava,nini}. Researchers can find loan agreements in exhibits to 10-K, 10-Q, and 8-K filings~\cite{nini}. Finance has learned once already that word lists built for general text misread 10\mbox{-}Ks until they were adapted~\cite{loughran2011}.

What does a raw agreement rate measure? Each extraction system's output is the target value, the answer the researcher's definition gives for that filing, plus an error in two parts. One part is shared by all systems, partly because they read the same documents under the same rules. The other part is specific to each system. When we compare two outputs, the target value and the shared part cancel. Raw agreement therefore measures how the systems differ from one another, not from the truth. The variable's prevalence and the firms it flags still carry the shared part. Two systems that both miss a threshold stated only in an exhibit agree, and both are wrong for a shared reason.

Because the systems share so much, the shared part can be large. Agreement can stay flat while the variable moves, for a simple reason. A rule that is inherited by all systems can act mainly through the shared part, which agreement cannot see. Not every inherited rule has this effect, as Section~\ref{sec:more} shows.

I audit a real archived extraction pipeline. In it, four open-weight configurations read identical selected text from a stratified sample of 15,600 10-K filings and flagged numerical covenant thresholds. Three results stand out.

\begin{figure}[t]
\centering
\begin{tikzpicture}
\begin{axis}[scale only axis,width=\columnwidth-4.3cm,height=3.7cm,xmin=-.4,xmax=3.55,ymin=-.6,ymax=9.9,y dir=reverse,
ytick={0,1,2,3,4.3,5.3,6.3,7.3,8.3,9.3},yticklabels={Gemma 4 26B (G26),Gemma 4 31B (G31),Nemotron 3 Super 120B (N120),Qwen 3.6 35B (Q35),G26/G31,G26/N120,G26/Q35,G31/N120,G31/Q35,N120/Q35},
tick label style={font=\footnotesize},label style={font=\footnotesize},xlabel={Change when debt exhibits count (pp)},
xmajorgrids=true,grid style={gray!20}]
\addplot[gray,dashed] coordinates{(0,-.6)(0,9.9)};
\addplot[gray!50] coordinates{(-.4,3.65)(3.55,3.65)};
\node[font=\footnotesize,anchor=east,blue!55!black] at (axis cs:2.3,1.5) {Prevalence};
\node[font=\footnotesize,anchor=west,brown!80!black] at (axis cs:0.55,6.8) {Disagreement};
\addplot+[only marks,mark=*,mark size=1.6pt,color=blue!55!black,mark options={fill=blue!55!black,draw=blue!55!black},error bars/.cd,x dir=both,x explicit]
table[x=x,y=y,x error=e]{
x y e
2.955515230 0 0.246926716
2.932145100 1 0.246173805
2.777914322 2 0.239241197
2.615470784 3 0.232447935
};
\addplot+[only marks,mark=*,mark size=1.6pt,color=brown!80!black,mark options={fill=brown!80!black,draw=brown!80!black},error bars/.cd,x dir=both,x explicit]
table[x=x,y=y,x error=e]{
x y e
0.063026885 4.3 0.144869388
0.296679871 5.3 0.146882254
-0.010767168 6.3 0.144328200
0.206122129 7.3 0.153436712
-0.038328220 8.3 0.148791678
0.163180270 9.3 0.146236210
};
\end{axis}
\end{tikzpicture}
\caption{Counting debt exhibits moves prevalence but barely moves disagreement. Each dot is the change, in percentage points (pp), when extractions from debt exhibits count in addition to the 10-K body, on the same 15,600 filings. Top: each system's prevalence, the weighted share of filings it flags. Bottom: each pair's disagreement, the weighted share of filings on which the two systems differ. Bars: approximate 95\% design-based confidence intervals.}
\label{fig:scope}
\end{figure}
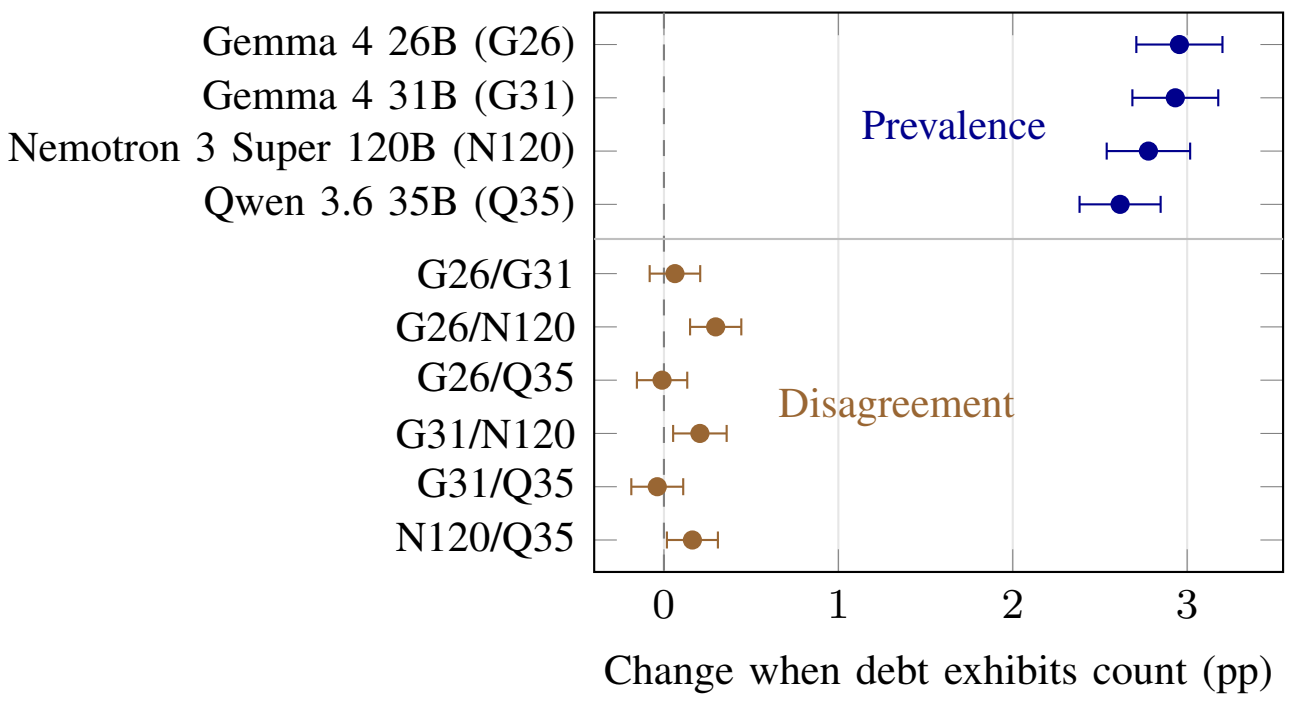

First, I hold every extraction fixed and change one rule: whether or not the extractions already made from a filing's debt exhibits count. Under the primary-only rule, which counts only the 10-K body, every system flags about one filing in five. A system's prevalence is the sampling-weighted share of filings it flags. Counting the exhibit extractions raises each system's prevalence by 2.6 to 3.0 percentage points, yet pairwise disagreement moves by at most 0.30 points (Fig.~\ref{fig:scope}). Two configurations I ran later on the same selected text behave similarly: counting the exhibits raises the prevalence of each by 2.8 points. Disagreement in the nine pairs that include at least one of them changes by $-0.03$ to $+0.35$ points.

Second, the rule changes a finding on a trend across time that a researcher would report. When only the 10-K body counts, the measured rise in covenant disclosure between 2000--2006 and 2020--2025 is about 13 points. However, when exhibits count as well, it is under 8 points, shedding about two fifths of the trend. Pairwise disagreement moves far less. In 2000--2006, where counting exhibits raises each system's prevalence by about 7 points, disagreement rises by 0.1 to 1.1 points. In later periods, it changes by at most 0.3 points.

Third, LLM readers outside the pipeline (Anthropic and OpenAI models) read a stratified sample of 420 source filings in full. Their reading is used to estimate how much agreement leaves unseen. Under the combined rule, where all four systems agreed on flagging a threshold, the readers find one in 138 of 140 sampled filings. Where all four agreed to flag none, they find one in 16 of 220, and in 50 more such filings read for a second rule. Among unanimous negatives, the share of filings that state a threshold (the false omission rate) is about 5\% under the researcher's definition and about 1\% if only plain maintenance tests count. Seen from the other side, these shared misses limit every system's recall. Under the researcher's definition, an estimated one in eight filings that state a threshold is flagged by none of the four, so no system's recall can exceed about 88\%. None of the 16 sampled shared misses is a clause that the systems had in front of them and simply overlooked. Instead, each traces to rules that all four systems inherit: which passages are selected, which quotes the validator accepts, which numbers are parsed, and which kinds of clause the prompt asks the models to record as thresholds.

I show why shared error cancels in pairwise agreement, derive an identity behind a short reporting protocol, and measure the shared error in an archived pipeline against readings of the full source filings.

The closest precedent finds that LLM scores of the same earnings calls diverge across models~\cite{boustanifar}. I document the opposite failure, flat agreement beside a moving variable, in covenant data of the kind LLMs already supply to research~\cite{covenantai}. In a companion study on the same filing sample, Gemma~4 26B flags current or past shocks in 34\% to 89\% of an 8,987-filing panel, depending on which sections and labels count~\cite{gun2026acl}. In a separate paper~\cite{gun2026gtrex}, two domain experts judged whether source text replacing 100 random LLM quotes kept their meaning. They disagreed on 17 of the 100, so human labels carry error too. When two LLMs judged the same 100, they agreed on 86, and on six of those both experts chose a different label~\cite{gun2026acl}. This paper ranks no system by accuracy and estimates no causal effect. Its subject is what an agreement rate can and cannot certify.

\section{Pipeline and Covenant Variable}\label{sec:data}
Our data started with 196,140 public SEC EDGAR 10-K filings from 2000 to 2025. I keep the earliest filing per registrant (CIK) and filing year, which leaves 192,469 original filings. I then exclude 6,950 filings from REITs, 19,878 from asset-backed issuers, and 45 with unusable content, leaving a sampling frame of 165,596.

Within each filing year, I rank the frame by the file size in bytes of the primary document (the 10-K body without its exhibits). I then split that year's filings into short, medium, and long thirds, with cutoffs depending on each year's distribution, as that varied significantly across the time window I studied. The long third starts at only 0.23~MB in 2000 and at 3.2~MB in 2025. From each of the 78 year-size cells (26 years times three size groups), I draw 200 filings at random, totaling 15,600 filings from 9,603 registrants. Each sampled filing has weight $w_i=N_h/200$, where $N_h$ is the number of frame filings in its year-size cell $h$.

I use four leading mid-sized open-weight models available when the pipeline ran. One of them is a dense model: Gemma~4 31B (\Gth{}) uses all of its 31B parameters for every token~\cite{gemmateam2026gemma4}. The remaining three models are mixture-of-experts models, in which a router sends each token through only a few of the model's experts. Gemma~4 26B A4B (\Gtw{}) has 26B parameters, of which about 4B are active for each token~\cite{gemmateam2026gemma4}. Qwen~3.6 35B A3B (\Qwe{}) has 35B, with about 3B active~\cite{qwen36_35b_a3b}. Nemotron~3 Super 120B A12B (\Nem{}) has 120B, with about 12B active~\cite{nvidia_nemotron_3_2025}. \Gtw{}, \Gth{}, and \Qwe{} ran with 8-bit weights and \Nem{} with 16-bit. All four ran in instruct mode. I call each configuration a system, because the unit that produces an answer in the pipeline is a base model together with its weight precision and its decoding settings.

All four systems read the same exact selected text. It comes from the 15,600 primary documents and from 3,553 material-contract (EX-10) and 957 debt-instrument (EX-4) exhibits attached to 2,195 of the filings. The pipeline retained these exhibits as debt contracts because terms such as borrower, lender, or trustee recur throughout them. Fixed keyword and pattern rules select 161,977 text passages, which I call windows. In 739 filings they select none. The windows are the same for every system and cover a median of 12.32\% of a document's characters across the 20,109 non-empty documents. Because every system reads the same selection, input differences cannot cause disagreement, but every system misses the evidence the rules leave out. I later ran two further systems on the same windows with the same prompt, schema, and validation rules. Both are mixture-of-experts models as well. GPT-OSS 120B (\Oss{})~\cite{openai2025gptoss120bgptoss20bmodel} has 117B parameters, with about 5.1B active for each token. It ran with its expert weights in 4-bit MXFP4. Qwen 3.8 Flash Next (\Qfl{})~\cite{qwen2026design,qwen38flashnext,qwen38flashnexthf} has 180B parameters in all. Its language model, the main processing part, holds 125B, of which about 6B are active for each token. The rest are a 51B n-gram embedding, a table of vectors for short token sequences that the model looks up rather than computes through, and a 4B module for multi-token prediction. \Qfl{} ran with 4-bit weights. I use these two only as a replication, in Sections~\ref{sec:results} and~\ref{sec:more}.\footnote{GPT-OSS has no non-reasoning mode, so it ran at its lowest reasoning effort. I also ran \Oss{} with 6-, 8-, and 16-bit weights and found no meaningful change on this task. \Qfl{} ran in instruct mode, like the first four.}

\emph{Prompt and schema.} Every system in this study receives the same exact system prompt and reads one window at a time. It returns one JSON object under a fixed schema with eight record arrays: facilities, financial covenants, reported ratios, compliance statements, covenant events, negative covenants, covenant definitions, and other thresholds. The prompt defines a financial-covenant record as ``a maintenance or incurrence test the borrower must satisfy under a debt instrument.'' Each such record holds the test's label, its direction (a maximum or a minimum), the threshold as written, any schedule of period-specific levels, the condition under which a springing covenant applies, a tag for the governing instrument, and a quote from the window. Clauses that resemble covenant tests but are not tests go to other arrays. Regulatory capital requirements, pricing grids, stock-exchange listing standards, covenants in leases or preferred stock, and hypothetical language go to the other-thresholds array. Restrictions on the borrower's actions, with any dollar or percentage basket, go to negative covenants. Each system ran once using stochastic decoding at its developer's recommended sampling settings (temperature 0.7 for the two Qwen models and 1.0 for the others).

\emph{Validation rules.} Three deterministic rules stand between a system's output and the indicator. First, the pipeline drops any record whose quote it cannot locate in the window, even after normalizing case, spacing, and quotation marks (the separate paper~\cite{gun2026gtrex} audits 10,000 of the stored quotes and their offsets). Second, a threshold counts as numerical only if the expression as written parses to a number, as ``4.00 to 1.00'' does. A threshold stated as a formula therefore stays as text and does not count. Third, an instrument filter moves a financial-covenant record to the other-thresholds array, and so out of the indicator, when the system tagged its governing instrument as something other than debt (a lease, regulatory capital, preferred stock or a charter, another contract, or a hypothetical). Each rule applies identically to every system, so errors from these rules are candidates for the common component of Section~\ref{sec:model}.

The target asks whether filing $i$'s archived package $\mathcal F_i$ explicitly discloses a numerical financial-covenant threshold on the debt of the registrant or its consolidated group. The package is the 10-K body with its included debt exhibits (those retained by the pipeline). Agreements incorporated by reference from earlier filings lie outside the package. A maximum leverage ratio covenant of 4.0 qualifies, but a reported leverage ratio of 2.6 on its own does not. Both maintenance and incurrence tests qualify. The borrower must meet a maintenance test at each testing date while the debt is outstanding, and an incurrence test before an action such as new borrowing or a dividend. Historical, future, and springing thresholds also qualify, as do a level stated as a formula, a trigger level, and a clean-down requirement. A trigger level is a stated level that, when crossed, brings a covenant into force or changes the creditor's rights. A clean-down requirement obliges the borrower to pay a revolving loan down to a stated level for a set period. Under the target, a level stated as a formula counts as numerical, although the second validation rule keeps it as text.

Loan principal, maturity, interest rates, pricing grids, fixed-dollar baskets, and cash collateral do not qualify. The variable says nothing about compliance or slack. This target is the researcher's, here mine, and in a further respect it is wider than what the indicator counts. Trigger levels, limits on dividends and other payouts tied to a financial measure, and clean-down requirements count under the target, but the prompt does not name them as financial-covenant thresholds. The prompt keeps the trigger of a springing covenant only as condition text, which the indicator does not read. A payout limit is an incurrence test under the target, but it also reads as a restriction on the borrower's actions, which the prompt sends to negative covenants. I return to this gap in Section~\ref{sec:sources}.

Let $m=1,\dots,M$ index the $M=4$ systems. Call the definition above $\theta$. It fixes which clauses, periods, entities, and debts qualify. The target value of filing $i$ is the answer obtained by applying this definition to the filing's package: $Y_i^*=g(\mathcal F_i;\theta)$, where $g$ stands for reading the package $\mathcal F_i$ under $\theta$. I call it the target value because it is fixed by the target, the researcher's definition, and not by what any system reports: it is correct relative to that definition, and a different definition can give a different target value. It equals 1 if the package discloses a qualifying threshold, 0 if it does not, and $\bot$ if the case is indeterminate. The star marks it as the value the systems are meant to recover, as distinct from their own indicators (defined next), which may be wrong. For system $m$, $P_{im}$ indicates whether at least one numerical financial-covenant record from the primary document survives the validation rules, and $E_{im}$ indicates the same for the included debt exhibits. The combined filing indicator is
\begin{equation}
C_{im}=\max(P_{im},E_{im}),\qquad C_{im}-P_{im}\in\{0,1\}.
\label{eq:scope}
\end{equation}
Changing which documents feed the indicator changes neither $\theta$ nor $Y_i^*$. In one sampled filing, a credit agreement filed as an exhibit states that ``the Company will not permit the Leverage Ratio as of the last day of any fiscal quarter to exceed 4.00 to 1.00.'' The body of the same 10-K states no level. This exhibit-only threshold gives the filing $P_{im}=0$ and $E_{im}=1$ for every system.

Rates are $w_i$-weighted means over the sampled filings and estimate shares of the frame. A group's share of frame weight is thus the share of the 165,596 frame filings it represents. Intervals for rates and changes computed on all 15,600 sampled filings are approximate 95\% design-based confidence intervals that hold the extractions fixed. They follow the sampling design, which draws filings without replacement within each year-size cell and independently across cells: each is $\pm1.96$ standard errors of the stratified mean with a finite-population correction, computed on per-filing differences for a change. They therefore describe the frame, so there is no need to cluster by registrant. Intervals from the smaller samples of Section~\ref{sec:sources} follow those samples' designs. They cover sampling error only, not labeling or run-to-run error. An interval that contains zero does not show equivalence.

\section{Agreement and the Common Component}\label{sec:model}
In its simplest form, each system's error can be decomposed into two parts. One of these parts is specific to the system: run-to-run noise, the chosen LLM's own error, and its interaction with the rest of the pipeline. The other one is shared by every system, whether a deterministic pipeline rule imposes it on all alike or the systems themselves produce it. Raw agreement sees only the first. Let $\rho$ denote such a rule. The first rule I study is document scope: the primary-only rule counts only the 10-K body, and the combined rule also counts the debt exhibits. Write $Z_{im}(\rho)$ for the output of system $m$ on filing $i$ under $\rho$, so that $Z_{im}$ is $P_{im}$ under the primary-only rule and $C_{im}$ under the combined rule. Each output is the target value plus these two parts: a common component $c_i(\rho)$, the part that enters the output of all $M$ systems, and a component $e_{im}$ specific to system $m$, which may also depend on $\rho$. For an extracted amount, such as a dollar figure, the three terms add. For a yes/no indicator, such as the covenant flag studied here, they add modulo two, written $\oplus$, so a component equal to one flips the answer:
\par\noindent{\small\abovedisplayshortskip=\abovedisplayskip\begin{equation}
Z_{im}(\rho)=Y_i^*+c_i(\rho)+e_{im},\qquad Z_{im}(\rho)=Y_i^*\oplus c_i(\rho)\oplus e_{im}.
\label{eq:model}
\end{equation}}%
For amounts, $Y_i^*$ is the amount under the target definition. Where the target is indeterminate, any placeholder for $Y_i^*$ serves, since it cancels in every contrast. I drop $(\rho)$ when the rule is clear. Because $c_i(\rho)$ is common to the $M$ systems, it is defined relative to that set of systems.

Both forms are identities for any split of each system's error into a part shared by all systems and a system-specific part. The lemma below applies to all such decompositions. Where all systems return the same value, every split gives all $e_{im}$ the same value, because their pairwise contrasts are zero. I normalize that value to zero, so that the whole error is common, and $c_i(\rho)$ is nonzero (for indicators, equal to one) exactly when the shared value is wrong. On mixed filings there is no privileged split. The substance of the model is that a rule every system inherits can act through $c_i(\rho)$, not only through $e_{im}$, and that $c_i(\rho)$ may depend on the filing, including on $Y_i^*$. For instance, if a filing's only threshold sits in an exhibit that the rule excludes and every system returns zero, then $c_i=1$.

I assess agreement between two systems. Let $\mathbb E[\cdot]$ denote an average over the filing frame, which I estimate by the $w_i$-weighted average over the sampled filings. Statistics based solely on the systems' outputs use every filing. Statements about error against the target concern filings with a determinate target. For indicators, the prevalence of system $m$ is $p_m=\mathbb E[Z_{im}]$. For systems $a$ and $b$, the pairwise disagreement is $D_{ab}=\mathbb E[\ind(Z_{ia}\ne Z_{ib})]$ and the positive union (the share flagged by either system) is $U_{ab}=\mathbb E[\max(Z_{ia},Z_{ib})]$. Conditional disagreement within the union is $q_{ab}=D_{ab}/U_{ab}$ for $U_{ab}>0$, and $1-q_{ab}$ is the weighted Jaccard similarity of the two positive sets. Four-way agreement, or unanimity, is $\mathbb E[\ind(Z_{i1}=\dots=Z_{iM})]$.

\begin{lemma}[Agreement is blind to the common component]\phantomsection\label{lem:main}%
Under \eqref{eq:model}, $Z_{ia}-Z_{ib}=e_{ia}-e_{ib}$ for amounts and $\ind(Z_{ia}\ne Z_{ib})=e_{ia}\oplus e_{ib}$ for indicators. Consequently, every statistic that depends on the outputs only through such pairwise contrasts (the rates of exact equality and of equality within a fixed absolute tolerance, the quantiles of $|Z_{ia}-Z_{ib}|$, unanimity, and $D_{ab}$) is a function of the $e_{im}$ alone and invariant to $c_i(\rho)$ and to $Y_i^*$. Agreement within a relative tolerance $0<t<2$ is invariant to a common rescaling of all outputs but not to a shared additive shift. The quantities a researcher uses the variable for can change with $c_i(\rho)$: the error against the target ($Z_{im}-Y_i^*=c_i(\rho)+e_{im}$ or $\ind(Z_{im}\ne Y_i^*)=c_i(\rho)\oplus e_{im}$), the prevalence $p_m(\rho)$, the firms the variable flags, and regression coefficients that use $Z_{im}$. These can change even when $c_i(\rho)$ is independent of $Y_i^*$, the regressors, and the outcome, as under misclassification, and can be biased either way when it covaries with them.
\end{lemma}
\begin{IEEEproof}
Subtracting the outputs of two systems, or adding them modulo two, cancels $Y_i^*$ and $c_i(\rho)$ in \eqref{eq:model}. A relative tolerance, $|Z_{ia}-Z_{ib}|\le t\max(|Z_{ia}|,|Z_{ib}|)$, is unchanged when every output is multiplied by the same nonzero constant. A shared shift can never move the left-hand side but can move the right-hand side: if $Z_{ia}\ne Z_{ib}$, a shift by $-(Z_{ia}+Z_{ib})/2$ makes the pair fail every tolerance $t<2$, while a large enough shift makes it pass. Each quantity in the last claim depends on the outputs themselves, not only on their contrasts. Holding $Y_i^*$ and every $e_{im}$ fixed, a change in $c_i(\rho)$ on filings of positive weight changes every $Z_{im}$ on them and leaves every pairwise contrast unchanged. A rule that also changes the $e_{im}$ can change the contrasts as well. For a regression example, set every $e_{im}=0$, let $Y_i^*$ be a binary regressor with mean one half, let the outcome equal $Y_i^*$, and let $c_i=1$ with probability $r$, independently of both. The population slope on the observed indicator is then $1-2r$ rather than 1. For an amount regressed on $X_i$, again with $e_{im}=0$, the slope shifts by $\mathrm{Cov}(c_i,X_i)/\mathrm{Var}(X_i)$.
\end{IEEEproof}

\begin{corollary}[Monotone rule change]\phantomsection\label{cor:scope}%
Let the rule change so that no system's output can fall from $1$ to $0$. This holds when switching from primary-only to combined, where each $Z_{im}$ changes by $C_{im}-P_{im}\in\{0,1\}$ as in \eqref{eq:scope}, or when lifting the instrument filter of Section~\ref{sec:more}. For a pair of systems $a$ and $b$, take a filing on which both move from negative to positive. In the model applied to that pair, we can represent this move as a change in the pair's shared component alone. It raises $p_a$, $p_b$, and $U_{ab}$ by the filing's share of frame weight and leaves $D_{ab}$ unchanged. Taken alone, it therefore lowers $q_{ab}$ when $D_{ab}>0$. Code the pair's state $(Z_{ia},Z_{ib})$ as $00$, $11$, or $1$ for either one-sided state, $01$ or $10$. Because no output falls, every filing that changes state makes one of three moves: $00\to11$ (a shared move), $00\to1$ (a created disagreement), or $1\to11$ (a resolved disagreement). Write $\pi_{00\to11}$, $\pi_{00\to1}$, and $\pi_{1\to11}$ for their shares of frame weight. Then $\Delta D_{ab}=\pi_{00\to1}-\pi_{1\to11}$ and $\Delta U_{ab}=\pi_{00\to11}+\pi_{00\to1}$, while $p_a$ and $p_b$ each rise by at least $\pi_{00\to11}$. These are accounting identities for the two outputs and hold without any model. Shared moves therefore leave $D_{ab}$ untouched, whatever their weight may be. When $U_{ab}>0$ before the rule change, conditional disagreement $q_{ab}$ falls if and only if $\Delta D_{ab}<q_{ab}\,\Delta U_{ab}$, with $q_{ab}$ evaluated before the change. Chance-corrected coefficients such as Cohen's $\kappa$ and Gwet's AC1 equal $1-D_{ab}/(1-p_e)$ for a chance term $p_e<1$ that depends on $p_a$ and $p_b$, so they can move too. Shared moves leave $D_{ab}$ and $p_a-p_b$ unchanged and increase the mean prevalence $(p_a+p_b)/2$. When $D_{ab}>0$ and that mean stays below one half, shared moves therefore raise $\kappa$ and lower AC1. One-sided moves can add to or offset these changes. A flat $D_{ab}$ beside a rising prevalence signals shared or offsetting moves, not stability.
\end{corollary}

The model does not introduce a coefficient and does not rank systems. It tells us which component each reported statistic responds to. On unanimous filings the normalization sets every $e_{im}$ to zero, so only reading the source can reveal $c_i(\rho)$. On mixed filings such a reading gives each system's total error, but the split between $c_i(\rho)$ and $e_{im}$ remains a convention. Reading a weighted sample of unanimous filings therefore estimates the share of frame weight on which all systems agree and are wrong (Section~\ref{sec:sources}).

\section{The Document-Scope Rule}\label{sec:results}
\begin{table}[t]\centering\footnotesize
\caption{Pairwise Disagreement and Transition Ledger by Document Scope (Weighted \%, 15,600 Filings)}
\label{tab:ledger}
\setlength{\tabcolsep}{3pt}
\begin{tabular}{lccccccc}\toprule
&\multicolumn{2}{c}{$D$}&\multicolumn{3}{c}{$\pi$, Prim.\ to Comb.}&\multicolumn{2}{c}{$q=D/U$}\\\cmidrule(lr){2-3}\cmidrule(lr){4-6}\cmidrule(lr){7-8}
Pair&Prim.&Comb.&$00\to11$&$00\to1$&$1\to11$&Prim.&Comb.\\\midrule
G26/G31 & 5.32 & 5.38 & 2.47 & 0.51 & 0.45 & 22.92 & 20.56 \\
G26/N120 & 5.74 & 6.04 & 2.38 & 0.64 & 0.34 & 25.11 & 23.33 \\
G26/Q35 & 6.12 & 6.11 & 2.32 & 0.46 & 0.48 & 25.34 & 22.68 \\
G31/N120 & 5.94 & 6.14 & 2.32 & 0.64 & 0.43 & 25.26 & 23.22 \\
G31/Q35 & 6.18 & 6.14 & 2.27 & 0.49 & 0.53 & 24.98 & 22.34 \\
N120/Q35 & 6.01 & 6.17 & 2.20 & 0.58 & 0.41 & 24.94 & 22.97 \\
\bottomrule\end{tabular}\\[3pt]
\parbox{\columnwidth}{\footnotesize Prim.: primary-only rule (10-K body only). Comb.: combined rule (body plus retained debt exhibits). $D$: share of filings on which the two systems differ. $U$: share flagged by either. $\pi$: frame weight of shared moves ($00\to11$), newly one-sided moves ($00\to1$), and resolved disagreements ($1\to11$), where $00$ means neither system flags the filing, $11$ both, and $1$ exactly one. Prevalence $p_m$, Prim./Comb.: \Gtw{} 20.00/22.95, \Gth{} 21.09/24.02, \Nem{} 19.97/22.75, \Qwe{} 22.18/24.80. Changes recomputed from the rounded entries can differ by 0.01.}
\end{table}

Counting debt exhibits moves the covenant variable while leaving agreement nearly where it was. Under the base case, the primary-only rule, the four systems put weighted covenant prevalence at about 20\% to 22\%. When extractions from retained debt exhibits also count, prevalence rises in every system by 2.6--3.0 percentage points (Fig.~\ref{fig:scope}, Table~\ref{tab:ledger}). Unconditional pairwise disagreement $D$ changes by only $-0.04$ to $+0.30$ points across the six pairs, and $q$ falls by 1.8--2.7 points (Table~\ref{tab:ledger}). Judged by agreement among flagged filings, a larger and different covenant sample even looks more reliable.

Table~\ref{tab:ledger} gives the accounting behind Corollary~\ref{cor:scope}. In every pair, 2.2\%--2.5\% of frame weight moves from both-negative to both-positive ($\pi_{00\to11}$). These shared moves raise prevalences as well as the union $U$, but leave $D$ untouched. Only one-sided moves can change $D$. Newly one-sided moves carry about 0.5\%--0.6\% of frame weight ($\pi_{00\to1}$), whereas resolved one-sided disagreements carry about 0.3\%--0.5\% ($\pi_{1\to11}$). In the three pairs without \Nem{}, the two nearly offset each other, and $D$ changes by $-0.04$ to $+0.06$ points. In the three pairs with \Nem{}, newly one-sided moves outweigh resolved disagreements, and $D$ rises by 0.16--0.30 points, with 95\% intervals that exclude zero (Fig.~\ref{fig:scope}). The pattern is not confined to one model family. In each of the five cross-family pairs tested, as in the only same-family pair (\Gtw{}/\Gth{}), shared moves outweigh newly one-sided moves by more than three to one, and $q$ falls while $D$ barely moves.

The same accounting holds for the four systems as a group. Of the 15,600 filings, 601 change their four-system pattern. Of these 601 filings, 310 move from all four negative to all four positive. They carry 1.92\% of frame weight and belong to the shared moves of every pair. The rest move from all negative to mixed (150), from mixed to all positive (113), or from one mixed pattern to another (28). Weighted, the all-positive share rises from 15.36\% to 17.99\%, yet four-way agreement slips only from 89.17\% to 88.94\%. The shared moves leave unanimity untouched, as Lemma~\ref{lem:main} says they must.

Take the \Gtw{}/\Nem{} pair, whose disagreement moves the most. \Gtw{} adds 476 positives and \Nem{} adds 448. In 384 filings both systems switch together, so $D$ cannot see them. Disagreement grows only by the 102 filings that become one-sided, less the 54 that resolve. Weighted, $D$ rises by 0.30 points (95\% interval $[0.15,0.44]$), a change that survives a Holm adjustment across the six pairs. The rise for \Gth{}/\Nem{} also narrowly survives (adjusted $p=0.04$).

\emph{Does it hold for other models?} For the two I added, yes. They read the same windows under the same prompt, schema, and validation rules, so they vary the model configuration while the pipeline stays fixed. When exhibits are counted, prevalence rises by 2.8 points for both \Oss{} (from 20.0\%) and \Qfl{} (from 24.8\%). Across the nine pairs that include a new system, $D$ changes by $-0.03$ to $+0.35$ points, and $q$ falls in all nine. As \Nem{} does among the original six pairs, one system accounts for the largest changes: $D$ rises by 0.24--0.35 points in \Oss{}'s pairs with \Gtw{}, \Gth{}, and \Nem{}. Together with \Gtw{}/\Nem{}, those three are the only pairs that survive a Holm adjustment across all 15 pairs. Even so, each rise in $D$ is less than a seventh of either system's rise in prevalence. In 270 of the 310 four-way shared moves, all six systems move from negative to positive, and weighted six-way agreement slips only from 85.70\% to 85.33\%.

\emph{Is it run-to-run noise?} For \Gtw{}, not the shared moves and only partly the disagreement. Each system ran once with stochastic decoding. I therefore re-ran \Gtw{} with identical inputs and settings on the 310 four-way shared moves and on a random 500 of the remaining sampled filings, excluding the 120 with both EX-4 and EX-10 exhibits, which I re-ran as a separate group. Although I selected the 310 on its original output, the re-run reproduces \Gtw{}'s move on 308 of them. On the 500, unweighted, the re-run disagrees with the original run on 2.8\% of filings under both document rules, and with the other three systems on 5.3\% on average. Its run-to-run variation is therefore real and about half the size of its mean disagreement with the other three systems (difference 2.5 points, paired bootstrap 95\% interval $[0.6,4.4]$). Each pairwise $D$ also carries the other system's run-to-run variation, so this difference approximates the stable part of $D$ only if the other systems vary from run to run about as much as \Gtw{} does.

\emph{Would a better statistic see it?} Standard ones would not see it. Their disagreement is there even for the sign of the change. Cohen's $\kappa$ rises in all of the six pairs, by 0.008--0.016, and four-system Fleiss $\kappa$ rises from 0.8215 to 0.8339. Conversely, Gwet's AC1 falls in all six pairs, by 0.004--0.009, and its four-system value falls from 0.9123 to 0.9062. Both chance terms depend on prevalence~\cite{feinstein,gwet}. As prevalence moves toward one half, the chance term $p_e$ moves by about 0.03 in every pair, falling under $\kappa$ and rising under AC1. A change in $D$ moves both coefficients the same way, so the split in sign comes from the chance terms alone. A reported reliability coefficient therefore reflects the document-scope rule as well as the systems.

A latent-class model does not do any better either. It sees only the outputs too. By \eqref{eq:model}, a filing with target $Y_i^*$ and common error $c_i$ yields the same outputs as one with target $Y_i^*\oplus c_i$ and no common error. Hence, the outputs alone cannot separate $c_i$ from $Y_i^*$. A Dawid--Skene model~\cite{dawid} fitted with the sampling weights to the four indicators reports high sensitivity (0.87--0.93) and specificity (0.98--0.99) for every system under both rules. Yet it reads a pure change in which documents count as a rise in true prevalence, from about 22\% to about 25\%, and it assigns a unanimous negative a posterior probability below 0.0001 of stating a threshold. It also fits poorly (weighted $G^2$, with the 15,600 filings as sample size, of 444 under the primary-only rule and 467 under the combined rule, each on 6 degrees of freedom). Every pair is positive together more often than the fitted model predicts, consistent with some filings being hard for all four systems. The misfit does not measure $c_i$: to the model, an error that all four systems share looks like the true value.

\section{Reading the Sources}\label{sec:sources}
Since agreement is incapable of telling us whether an answer shared by all four systems is right, I went back to the sources. I split the 15,600 filings into the six groups of Table~\ref{tab:audit}, which refine the filings' four-system pattern under the combined rule. From each one of those groups, I drew a random sample. Two frontier closed-weight LLM readers, from Anthropic and OpenAI, read each sampled filing outside the 310 shared moves. Neither is an extractor, and they read the filing in full: the 10-K body and every included debt exhibit, not the selected windows. (The 100 filings sampled from the 310 shared moves were read in October 2026 by earlier Anthropic and OpenAI models, with the Anthropic reader above also reading 30 of them. In November 2026, the OpenAI reader re-read all 100, and the Anthropic reader read the disputed 10-K bodies.) Working separately, each reader answered two questions: whether the 10-K body states a qualifying threshold and whether an included debt exhibit does. Each reader also quoted the clause it based its answers on.

The readers of each filing agreed on its label in 410 of the 420 filings. I manually adjudicated the other ten (eight outside the 310 shared moves and two among the 100 from them), the remaining differences on the body or the exhibits, and the boundary cases of the definition. By the argument of Section~\ref{sec:model}, this agreement cannot certify the labels the two readers share. The readers differed from the extractors in two ways: they read the full filings rather than the windows, and they answered the target question directly. So they do not inherit the selection, validation, and prompt rules studied here.

As a small check on the readers, I also blind-labeled 40 clauses. Of these clauses, 30 had a settled answer: 23 that both readers counted and 7 look-alikes that do not qualify. The blind labels matched 24 of the 30, left 4 undecided (two of each kind), and disagreed on 2 (one of each kind). The other ten clauses came from the ten filings outside the 310 where the readers differed on the body or exhibits, including the eight with a filing-label disagreement. I adjudicated those filings afterward. My rulings match the blind labels on eight.

\begin{table}[t]\centering\footnotesize
\caption{Source Reading of 420 Filings by Four-System Pattern Under the Combined Rule}
\label{tab:audit}
\setlength{\tabcolsep}{3pt}
\begin{tabular}{lrrrc}\toprule
Four-system pattern&Filings&Read&Threshold&95\% limits (\%)\\\midrule
All no, no text selected & 739 & 40 & 0 & 0.0--8.5 \\
All no, debt exhibit attached & 833 & 60 & 11 & 9.8--30.0 \\
All no, no debt exhibit & 9,458 & 120 & 5 & 1.4--9.4 \\
Mixed & 1,736 & 60 & 48 & 67.9--89.1 \\
All yes, 310 shared moves & 310 & 100 & 98 & 93.9--99.4 \\
All yes, other & 2,524 & 40 & 40 & 91.3--100 \\\midrule
Total & 15,600 & 420 & 202 & \\
\bottomrule\end{tabular}\\[3pt]
\parbox{\columnwidth}{\footnotesize Filings: group size. Read: sampled filings the LLM readers of Section~\ref{sec:sources} read in full. Threshold: read filings whose 10-K body or included debt exhibit states a qualifying threshold. The two ``debt exhibit'' rows split the unanimous negatives with selected text. The 310 shared moves are the filings each of the four systems flags only when exhibits count. Limits: exact finite-population 95\% limits on the unweighted share of the group's filings that state a threshold, given the readers' labels. The 56 filings of Section~\ref{sec:more} add 50 unanimous negatives outside this sample (8 and 42 in the two ``debt exhibit'' rows), which the text counts as known.}
\end{table}

\emph{Shared positives hold up.} All 40 filings sampled from the unanimous positives outside the 310 carried a threshold. So did 98 of the 100 sampled from the 310, the filings that all four systems flag only once exhibits count. Of those 100, 92 state the threshold only in an exhibit, 6 state one in the body as well, and 2 are false positives. In the model, all 98 true positives have $c_i=1$ under the primary-only rule and $c_i=0$ under the combined rule. The 2 false positives move from $c_i=0$ to $c_i=1$. The indicator moved because of the document rule, which applies to every extractor alike, and agreement was silent about it.

\emph{Shared negatives are less safe.} Where the selection rules did not return any text for any of the systems, the readers found no threshold in any of the 40 sampled filings. Among the other unanimous negatives, however, they found a threshold in 11 of 60 filings with an included debt exhibit and in 5 of 120 without one. Fifty more unanimous negatives are among the 56 filings of Section~\ref{sec:more}. None of the 50 was in the Table~\ref{tab:audit} sample. The same readers read all 50 in full and found a threshold in every one. Counting these 50 as known and scaling each group's $w_i$-weighted sampled rate to the rest of that group's frame weight, I estimate the false omission rate of the unanimous negatives, the share of them that state a threshold, at 5.2\% under the researcher's definition (95\% Korn--Graubard interval 2.5\%--9.2\%, which allows for the skew of a rate built from few positives). None of the four indicators captures these thresholds. These missed filings carry an estimated 3.7\% of frame weight (95\% interval 1.8\%--6.6\%), comparable to the 1.9\% carried by the 310 shared moves. No agreement statistic can tell them from correct unanimous negatives, because the four outputs are the same in both.

Under the normalization of Section~\ref{sec:model}, every such error is common, so the reading puts about 3.7\% of frame weight on unanimous negatives with $c_i=1$ under the combined rule. False unanimous positives can only add to this shared error.

\emph{Why do all four systems miss the same filings?} I sorted the 16 sampled misses by the clause the readers quoted. In five of these, the clause lies outside the text selected for the systems, so a selection rule hid it from all four at once. In two, the clause is a net-worth test. Its level is a formula: a base amount plus a share of later net income. The prompt asks systems to record such thresholds as text, and the second validation rule counts only thresholds that parse to a number. In eight, the clause is of a kind that the target counts but the prompt does not ask the models to record as a financial-covenant threshold. Five of these state the level at which a springing covenant or another contractual consequence takes effect. For the three springing covenants, the filing gives no level for the covenant itself, and the prompt asks for the trigger only as condition text, which the indicator does not read. Two tie dividends or other restricted payments to a financial measure. One requires the revolving loan to be paid down to zero for thirty consecutive days each year.

Only one miss is a plain maintenance test with a stated number inside the selected text, a minimum interest coverage ratio of 2:1, and all four systems extracted it. The sentence spans a page break, each system's quote leaves out the page number and running header, and the first validation rule dropped all four records. All 16 sampled misses thus trace to rules that every system inherits. If I count only clauses that both readers classify as plain maintenance tests, wherever they sit and however their level is stated, the sampled shared misses fall to 4 of 60 and none of 120. All 50 known misses from Section~\ref{sec:more} are of this kind, each turned negative by the instrument filter. With them, the estimate is about 1\% of unanimous negatives. With none of 120 sampled, on the other hand, the exact 95\% upper limit for the remaining filings without a debt exhibit is still 3.0\%.

\emph{How much do the systems miss overall?} Where the systems disagree, the readers find a threshold in 48 of 60 filings. This means that the systems we have used have a tendency towards Type II error over Type I error in the case of disagreements. After I weight the six groups and count the 50 known misses, the readers' labels put about 31\% of frame weight on filings with a qualifying threshold. The four systems report about 23\%--25\% under the combined rule. About 3.7 of the missing points lie in filings that all of the four pass over: about one in eight filings that state a threshold is not flagged by any of the four, so no system's recall (the share of such filings it flags) can exceed about 88\%. Part of the gap reflects clauses the prompt does not ask the models to record as thresholds.

A separate check at the document level also points to shared error. Given the same windows, prompt, schema, and validation rules, the Anthropic reader above, a frontier closed-weight model outside the four, relabeled a stratified sample of 500 documents with selected text, blind to the four systems' outputs. It differed from the four systems on 2.8\% of the documents on which they are unanimous (95\% interval 0.9\%--4.6\%). In six of the nine sampled differences, it flagged a document that all four left negative. I weight this rate to the documents with selected text in the sampled filings, not to the frame. Because the check shares the windows, the prompt, and the validation rules, it cannot see misses that these common elements cause by themselves, such as a clause outside the selected text or a level stated as a formula, nor errors the relabeling model shares with the four. It can see a rule act on the systems' own judgments: in two of the six, the instrument filter, the third validation rule, had set aside thresholds that all four, or three of the four, tagged as governed by something other than debt. Some differences may be its own errors.

\emph{Agreement among decision models.} On the same 500 documents, I asked twelve models from nine developers (Table~\ref{tab:decision}) whether each of the 3,943 windows states that a debt covenant binds the company or a subsidiary as borrower or issuer. These decision models return a yes probability. A document counts as yes if any window reaches one half. This question is broader than the target, so I map the Anthropic reader's records from the check above to the same scope. Across the 66 pairs, disagreement runs from 2.7\% to 20.2\% of documents. Yet the share of documents on which both models of a pair differ from the reader is 2.4 to 10.4 times the product of their separate rates of differing from it, where independence gives one (median 5.5, stratified bootstrap 95\% interval 4.7--6.5). Shared model biases, hard documents, and errors of the one reader can each produce this overlap. All twelve agree on 67.8\% of documents, and the reader differs from their common answer on 2.4\% of those (95\% interval 1.1\%--3.8\%). Reading these 11 sampled documents, I find two covenants all twelve missed, seven reader errors, and two unclear cases. Correcting the seven reader errors leaves the median ratio at 5.6.

\begin{table}[t]\centering\footnotesize
\caption{Twelve Decision Models on 500 Documents (Weighted \%)}
\label{tab:decision}
\setlength{\tabcolsep}{3pt}
\begin{tabular}{llrrr}\toprule
Model&Developer&Yes&Differs&Closest\\\midrule
OpenJev 4B&AlexWortega&69.7&10.9&7.4\\
elephant-1&Base Labs&63.5&8.3&4.7\\
joey-1&Base Labs&57.9&13.9&8.8\\
Clef&Cloudflare&59.3&9.5&2.7\\
Clef Flash&Cloudflare&58.1&9.2&2.7\\
d1-2B&LiquidAI&74.1&19.5&15.8\\
OpenJev 27B&OpenJev&62.6&6.7&3.8\\
pplx-decider-flash&Perplexity&58.0&9.6&3.8\\
Jev 1.14&TypeSafe&60.1&7.5&3.8\\
d3-turbo&vLLM Sem.\ Router&58.0&9.9&3.0\\
d3-ultra&vLLM Sem.\ Router&57.9&10.6&3.0\\
LFM3-VLD 2B&webbrain-one&55.5&18.5&10.0\\
\bottomrule\end{tabular}\\[3pt]
\parbox{\columnwidth}{\footnotesize Models in order of developer, weighted as in the preceding check. Yes: share called yes. Differs: share differing from the reader, which says yes on 65.5\% (any financial or negative covenant, covenant event, or compliance statement other than one of no financial covenants). Closest: lowest disagreement with another model. LFM3-VLD 2B puts a generative head on the d1-2B backbone and is scored at its first token.}
\end{table}

\section{A Second Rule and a Downstream Consequence}\label{sec:more}
\emph{A second rule does move agreement.} The document rule is one of several rules that every system inherits, and the same pipeline holds another rule as well: the instrument filter, the third validation rule of Section~\ref{sec:data}. I call any financial-covenant record whose threshold parses to a number a numerical record. I turn the filter off under the primary-only rule, so that numerical records whose governing instrument the models tagged as something other than debt also count. Records the models placed in the other-thresholds array themselves still do not count. Primary-only prevalence then rises by 1.1 (\Qwe{}) to 3.9 points (\Gth{}). Pairwise disagreement falls by 1.7--2.9 points. Every 95\% interval excludes zero, and lifting the filter under the combined rule gives similar results. Since $D$ falls, the one-sided flags this change resolves must outweigh those it creates. That alone does not show which component the rule acts through, since shared moves leave $D$ untouched. This rule's transition ledger shows it: in every pair, more frame weight moves from one-sided to both-positive (2.7\%--3.3\%) than from both-negative to both-positive (0.5\%--2.1\%). A shared move can be represented as a change in the pair's common component alone, while a resolved move cannot: it requires a change in a system-specific error, whatever else changes with it. In this sense the rule acts largely through the system-specific component. The tag is each model's own judgment: counting filings already flagged, \Gtw{} tags the instrument of a numerical record as regulatory capital in the body of 1,233 of the 15,600 filings, and \Qwe{} in 266. Only the 56 filings discussed below were read for this rule. Some new flags may therefore be genuine regulatory-capital requirements rather than debt covenants.

Even so, the rule has a shared part. In 56 filings (0.36\% of frame weight), all four systems are negative with the filter and positive without it. The same two readers of Section~\ref{sec:sources} read all 56 in full, and both find a debt-covenant threshold in the body of every one, such as a minimum net worth or a debt-service coverage ratio under a credit agreement. In each filing, all four systems had extracted at least one numerical record and tagged its instrument as something other than debt. When I match quotes, the first clause either reader cites is one of these records for all four systems in 51 of the 56 filings. All four tagged that clause as regulatory capital in 44. One plausible cause of the mistagging is shared as well: the prompt's list of regulatory-capital terms names capital, leverage-ratio, and liquidity requirements, which loan covenants also impose. The list does not name coverage ratios, the test in about a quarter of the cited clauses, so it cannot be the whole cause. The two added systems split on these filings. \Oss{} moves on 47 of the 56, but \Qfl{} already counts 41 with the filter on, so all six move together on only 12. A common component is common to the systems compared. Whether a rule moves agreement therefore depends on which component it acts through. Since $D$ registers only one-sided moves, it cannot show the shared part of either rule. The transition ledger, however, can.

\begin{table}[t]\centering\footnotesize
\caption{Prevalence and Pairwise Disagreement by Period and Document Scope}
\label{tab:era}
\setlength{\tabcolsep}{3.5pt}
\begin{tabular}{lccccr}\toprule
&\multicolumn{2}{c}{Prevalence (\%)}&\multicolumn{2}{c}{$D$ (\%)}&\\\cmidrule(lr){2-3}\cmidrule(lr){4-5}
Filing years&Prim.&Comb.&Prim.&Comb.&Shared ($n$)\\\midrule
2000--2006 & 11.9--13.9 & 19.0--20.6 & 3.9--5.4 & 4.4--5.8 & 212 \\
2007--2012 & 17.8--20.4 & 19.3--21.8 & 4.9--5.9 & 5.0--6.1 & 42 \\
2013--2019 & 25.4--27.8 & 26.7--29.0 & 5.6--7.0 & 5.3--6.7 & 33 \\
2020--2025 & 25.3--27.3 & 26.7--28.4 & 6.9--7.7 & 6.8--7.6 & 23 \\
\bottomrule\end{tabular}\\[3pt]
\parbox{\columnwidth}{\footnotesize Weighted ranges across the four systems (prevalence) or six pairs ($D$). Prim. and Comb.: as in Table~\ref{tab:ledger}. Shared: number of the 310 shared moves of Table~\ref{tab:audit} in the period.}
\end{table}

\emph{Can the scope rule change a conclusion?} Consider a simple descriptive question: has disclosure of numerical covenant thresholds become more common? Table~\ref{tab:era} reports prevalence by period. When only the 10-K body counts, prevalence roughly doubles between 2000--2006 and 2020--2025, rising by 12.8--13.6 points across the four systems. When exhibits also count, it rises by 7.4--7.8 points. Counting exhibits therefore removes 5.4--6.0 points, about two fifths of the primary-only rise. For each system, the 95\% interval for that reduction excludes zero (for \Gth{}, the smallest reduction, $[4.6,6.2]$).

The filings that turn positive under the combined rule are largely concentrated in the early years, where the gap between the two rules is widest (Table~\ref{tab:era}). Of the 310 four-way shared moves, 212 fall in 2000--2006. Only filings with a retained debt exhibit can turn positive when exhibits count. The archived package includes one for 22\% of filings in that period, compared with 10\%--12\% in later periods. Retention explains only part of this concentration. Among filings with a retained exhibit, 30\%--33\% turn positive in 2000--2006 against 9\%--13\% in 2020--2025, depending on the system tested. In the early filings, the body carries a flagged threshold less often and the exhibit carries one more often.

Agreement gives little warning of this. In 2000--2006, counting exhibits raises each system's prevalence by 6.7--7.2 points while raising each pair's disagreement by merely 0.1--1.1 points. Over the same period $q$ falls by 4.8--9.2 points, so the flagged filings again look more consistent, not less. In later periods, prevalence rises by 1.1--1.7 points and $D$ moves by $-0.3$ to $+0.2$ points. Under either rule, the systems agree with one another about the trend, even though the two rules disagree. The combined rule matches the target's document scope, although the resulting indicator still misses qualifying thresholds. A study of covenant prevalence, or one that selects firms on a disclosed threshold, would therefore reach a different early-period sample and a different trend under each rule. Agreement would not show which rule applies.

\section{Implications and Limitations}\label{sec:disc}
Raw agreement measures how the systems' errors differ from one another, not the error they share (Lemma~\ref{lem:main}). A pipeline that sends only disagreements to manual review would accept about 89\% of filings under either document rule. Counting exhibits would raise the share it accepts as positive from about 15\% to about 18\%. None of the 310 filings that all four systems switch would reach a reviewer.

I therefore propose reporting four things alongside the agreement rate. First, name the document scope and the target. An indicator built solely from the 10-K body, such as one built from the sections and notes a filing-text repository supplies~\cite{codesso}, naturally misses every threshold stated only in a debt exhibit. State the researcher's target apart from the prompt's categories, because the two can differ. The trigger level of a springing covenant counts under the target, but the prompt keeps it only as condition text, which the indicator never reads (Section~\ref{sec:data}). Second, report each system's prevalence. When reporting a chance-corrected coefficient, report Cohen's $\kappa$ and Gwet's AC1 together, since a change in prevalence alone can push them in opposite directions, as it does here (Section~\ref{sec:results}).

Third, for every rule that can be changed without rerunning the systems, report the transition ledger of Table~\ref{tab:ledger}: the filings that two systems switch together, and the one-sided switches that open or close a disagreement. Shared moves change prevalence but never disagreement, so disagreement cannot show them. When a rule can also remove flags, add the moves from positive to negative.

Fourth, read a sample of the filings the systems agree on. No statistic computed from the outputs can reveal an error that every system makes. Here, the readers' labels imply that, of the filings all four systems leave negative under the combined rule, about 5\% state a qualifying threshold (Section~\ref{sec:sources}). The sample should cover unanimous negatives, unanimous positives and, for a frame-wide estimate, mixed filings, each drawn with known inclusion probabilities. Expert labels on such a sample then support a design-based correction~\cite{egami} or finite-population bounds.

\emph{Limitations.} The main evidence comes from one task, a filing-level indicator, in one archived pipeline. All six systems read the same windows under the same prompt and validation rules, and the decision models and their reference share those windows. The lemma holds for any pipeline and for extracted amounts, but the magnitudes reported here describe this exact pipeline. The four original systems differ in model family, size, weight precision, and sampling settings at once, so the comparisons cannot attribute a difference to any one of these. I measured run-to-run variation only for \Gtw{}. I cannot separate a model's own error from its interaction with the fixed prompt and windows. The covenant trends describe the systems' flags for thresholds disclosed in the 10-K body alone or in the body and the filing's retained debt exhibits. They exclude agreements incorporated by reference and covenants disclosed only in other documents, such as 8-K or 10-Q filings, and they do not show how often firms had covenants in force. The scope rule changes a descriptive trend. I do not estimate a covenant regression, so I do not document how these shared errors affect regression estimates.

The source readers themselves are LLMs. They read the full filings and answer the target question directly, which keeps them from inheriting the pipeline's rules. Nevertheless, their agreement with each other carries the same caveat as the extractors'. If both readers overlook a threshold that all four systems also miss, the filing looks correctly negative and the 5\% estimate may be too low. Conversely, if both count a clause that does not qualify, the estimate can be too high. My own labels check the readers in part: I blind-labeled 40 clauses and adjudicated the ten disputed filing labels, the remaining differences on the body or exhibits, and the boundary cases. The estimate also depends on the definition (Section~\ref{sec:sources}).

For the instrument filter, the readers checked only the 56 filings that all four systems switch, so some other new flags may be genuine regulatory-capital requirements (Section~\ref{sec:more}). Finally, holding the extractions and reader labels fixed, the intervals measure sampling error only. The limits in Table~\ref{tab:audit} and the 3.0\% upper limit of Section~\ref{sec:sources} are exact for unweighted group shares, and the other intervals are approximate. I apply Holm adjustments across the pairwise tests of Section~\ref{sec:results} but make no adjustment for exploratory choices.

\section{Concluding Remarks}\label{sec:conc}
Although agreement among extractors is certainly useful, by itself it cannot certify the accuracy of a variable. In a real archived pipeline, counting extractions from debt exhibits raised the covenant prevalence of each of the four systems by 2.6 to 3.0 percentage points, while pairwise disagreement among them moved by at most 0.30 points. The same rule removes about two fifths of the measured rise in prevalence since the early 2000s. Reading a stratified sample of the source filings in full shows the same blind spot from the other side. On filings where all four systems agree under the combined rule that no threshold exists, LLM readers outside the pipeline put the share of such filings that state one, the false omission rate, at about 5\% under the researcher's definition, or about 1\% if only plain maintenance tests count. Under the researcher's definition, about one in eight filings that state a threshold is missed by all four, so no system's recall could exceed about 88\%. All 16 sampled misses trace to rules all four systems inherit: text the windows left out, quotes the validator rejected, formula-based thresholds the number rule drops, or clauses the prompt did not ask the models to record as thresholds.

A second inherited rule, a filter on thresholds the systems tag as governed by something other than debt, did lower the disagreement when I lifted it. Yet lifting it for the 10-K body also turns all four systems from negative to positive in 56 filings, and the readers find a debt-covenant threshold in the body of every single one. The reason agreement misses these shared errors is general. Raw agreement is a contrast between systems, and a contrast cancels whatever the systems share: both the error a common rule induces in all of them and the truth they were meant to recover. Naming the document scope, stating the target apart from the prompt's categories, reporting prevalence beside every agreement rate and the transitions under each rule that can be changed without rerunning the systems, and reading a sample of unanimous filings require only a modest change in practice. Together these steps let a researcher see the error the systems share and estimate it on the filings where they agree, before an LLM-extracted financial variable is used.

\section*{Acknowledgment}
I am forever grateful to Teoman Kadıoğlu for his guidance.
I also thank Mingjun Sun and Ian Smith for their help.

Large language models are central to the empirical analysis and are
described in the sections that use their outputs. Models from Anthropic,
Google, Z.ai, and OpenAI also assisted with research design, coding,
and checks, and, along with the Finference Prose model~\cite{gun2026prose}, helped draft and revise the text.

\balance

\end{document}